\RequirePackage{snapshot}
\documentclass[review,preprint,oneside,11pt]{elsarticle} 
\biboptions{sort&compress} \pdfoutput=1
\usepackage{graphicx}
\graphicspath{{./figs/}, {./figs/Biography/}}
\DeclareGraphicsExtensions{.pdf,.jpeg,.png}

\usepackage[cmex10]{amsmath}
\usepackage{mathtools}
\usepackage{amssymb}
\usepackage{algorithmicx}
\usepackage[ruled]{algorithm}
\usepackage[noend]{algpseudocode}

\usepackage{array}

\usepackage[hyphens]{url}

\usepackage{multirow}

\usepackage{float}

\usepackage{ulem}
 \usepackage{fixltx2e}
\usepackage{color}
\usepackage{soul}

\usepackage{colortbl}

\usepackage{afterpage} 
\usepackage{setspace}

\usepackage{xfrac}

\usepackage{picins} 
\usepackage{bm}

\usepackage{arydshln}

\usepackage{caption}

\usepackage{spreadtab}
\STautoround*{2} 
\usepackage{numprint}
\npthousandsep{,}\npthousandthpartsep{}\npdecimalsign{.}

\usepackage[strings]{underscore}

\usepackage{tabularx}

\usepackage[export]{adjustbox}
\usepackage{relsize}

\usepackage{comment}

\usepackage{enumitem}
\newcommand\litem[1]{\item{\bfseries#1.\space}}

\usepackage[export]{adjustbox}
\newcommand{\twincludegraphics}[2][]{   \includegraphics[
     width=\dimexpr\textwidth-2\fboxrule,
     frame={\fboxrule},
     #1]{#2}}

\usepackage{fnpos}

\def\por1{\partial}

 \DeclareMathOperator*{\argmin}{\,argmin} 

\makeatletter
  \newcommand\tinyv{\@setfontsize\tinyv{5pt}{7}}
\makeatother

\allowdisplaybreaks[1]

\makeatletter
\let\ftype@table\ftype@figure
\let\ftype@algorithm\ftype@figure
\makeatother

\newcommand{\LineComment}[1]{\State \(\triangleright\) \textit{#1}\hfill}
\algblock{ParFor}{EndParFor}
\algnewcommand\algorithmicparfor{\textbf{parfor}}
\algnewcommand\algorithmicpardo{\textbf{do}}
\algnewcommand\algorithmicendparfor{\textbf{end\ parfor}}
\algrenewtext{ParFor}[1]{\algorithmicparfor\ #1\ \algorithmicpardo}
\algrenewtext{EndParFor}{\algorithmicendparfor}
\makeatletter
\ifthenelse{\equal{\ALG@noend}{t}}   {   \algtext*{EndParFor}   }{}\makeatother

\newcolumntype{M}{>{\centering\arraybackslash}m{\dimexpr0.50\linewidth-2\tabcolsep}}

\newcommand{\mytitletile}{Modified Hausdorff Fractal Dimension (MHFD)\\
{\scriptsize Working Paper WP-RFM-15-02, (version: 150507)}}

\newcommand{\myauthorauthor}{
\author[syncets]{Reza~Farrahi~Moghaddam\corref{cor1}} 
\ead{imriss@ieee.org}
\ead[url]{http://ca.linkedin.com/in/rezafm}

\author[syncets]{Mohamed~Cheriet}
\ead{mohamed.cheriet@etsmtl.ca}

\cortext[cor1]{Corresponding author: Reza Farrahi Moghaddam}

\address[syncets]{\small Synchromedia Lab, \'Ecole de technologie sup\'erieure (ETS), University of Quebec (UduQ), Montreal, QC, Canada}

} 
\newcommand{\myabstractabstract}{
\begin{abstract}
The Hausdorff fractal dimension has been a fast-to-calculate method to estimate complexity of fractal shapes. In this work, a modified version of this fractal dimension is presented in order to make it more robust when applied in estimating complexity of non-fractal images. The modified Hausdorff fractal dimension stands on two features that weaken the requirement of presence of a shape and also reduce the impact of the noise possibly presented in the input image. The new algorithm has been evaluated on a set of  images of different character with promising performance.
\end{abstract}
}

\newcommand{\mykeywordskeywords}{
\begin{keyword}
Complexity \sep Fractal Dimension \sep Hausdorff Fractal Dimension \sep Illustrated Manuscripts
\end{keyword}
}

\begin{document}

\begin{frontmatter}

\title{\mytitletile}

\myauthorauthor

\myabstractabstract

\mykeywordskeywords

\end{frontmatter}

\section{Introduction}
\label{sec:intro}
The scale parameter plays an important role in image processing and understanding \cite{Farrahi2010c,Farrahi2012d}. However, it seems that the complexity parameter could be leveraged in order to improve the image processing methods. In many other fields, fractals have been extensively used to represent and also to compress patterns \cite{Iano2006}. To differentiate between fractals based on their degree of complexity without requiring a full model, a parameter called the Fractal Dimension (FD) has been commonly used \cite{Mandelbrot1967,Niemeyer1984,Mandelbrot1985,Zmeskal2003,Risovic2008,Soille1996}. In particular, the Hausdorff Fractal Dimension (HFD)\footnote{The HFD is a {\em divider} approach to estimating a fractal dimension \cite{Hausdorff1918,Carr1991}. It has been also referred to as the Hausdorff-Besicovitch dimension or the Richardson dimension.} \cite{Hausdorff1918} has been used in many applications. A simpler form of the Hausdorff fractal dimension is the Minkowski-Bouligand dimension or the Box-Counting Dimension (BCD) \cite{Dubuc1989,Kline1939}. In this work, from here on, we use the BCD's definition as that of the HFD.

Although the HFD measure is practical and fast to calculate, we argue that it could be improved by considering two features. Although these features will be discussed in greater details in Section \ref{sec_Modified_Hausdorff_Fractal_Dimension_MHFD}, they are briefly mentioned here. The first feature promotes incorporating `non'-object data in the calculations. It seems that this has been implicitly considered in the definition of the HFD for {\em fractals}, this could be easily violated when applying the HFD to estimate the complexity of images that are not by nature fractals. The second proposed feature is a scholastic process to rule out as much as possible of those boxes that are noise-related. Again although this feature may be of no-impact in the case of true fractals, it could improve the estimations for the degraded, noisy actual images. 

The paper is organized as follows. In Section \ref{sec_notations}, some of basic notations are defined. Then, the baseline definition of the HFD is presented in Section \ref{sec_Hausdorff_Fractal_Dimensio_HFD}. This is followed by the proposed definition of a Modified HFD (MHFD) in Section \ref{sec_Modified_Hausdorff_Fractal_Dimension_MHFD}. The illustrative examples are provided in Section \ref{sec_Illustrative_Examples}. Finally, the conclusions are presented in Section \ref{sec_Conclusions}.

\section{General Notation}
\label{sec_notations}
In this section, the notation used in the following sections is presented.
\begin{itemize}
\item{$I$}: The observed image (or patch): 
$$I=\big(I_{i,j}\big)_{i=1,j=1}^{n,m}=I_{(n,m)},$$
where $m$ and $n$ are the sizes of the image, and $I_{i,j}$ is an image pixel value at pixel position $\big(i,j\big)$. In this work, it is assumed that the image pixel values are binary: $I_{i,j} \in \big\{0, 1\big\}$, where $1$ is the value of an `object' pixel. For the purpose of reducing the ink use in print, the images may be shown in either BW01 or BW10 protocols\footnote{The BW01 protocol means that the $0$ pixels are shown in black while the $1$ pixels are shown in white \cite{Farrahi2010}.} \cite{Farrahi2010} depending on the ratio of the object to non-object pixels.
\item{$D_I$}: The fractal dimension of an image $I$.
\item{$B\big(i,j,w\big)$}: A square `box' or patch at the pixel position $\big(i,j\big)$ and with the patch width $w$. A particialut pixel $(k,l)$ in $B\big(i,j,w\big)$ is dented $B\big(i,j,w\big)_{k,l}$. In terms of the notations in \cite{Farrahi2012d}, a box $B\big(i,j,w\big)$ could be approximated to a patch $P_{(i+[w/2],j+[w/2]),[w/2],\infty}$.
\end{itemize}

\begin{algorithm}[!htbp]
\small
\caption{Calculate the HFD using the discrete $S$-tuples ${\cal S}_I$ and ${\cal N}_I$.}
\label{alg_HFD_1}
\begin{algorithmic}[1]
\Procedure{$\text{HFD}_I = \mathbf{HFDCalculate}$}{$I_{n,m}$}
	\LineComment{\emph{First, calculate ${\cal S}_I$ and ${\cal N}_I$}}
	\State $S \gets \max \left(\log_2 m, \log_2 n\right)$
	\ParFor {$s = 0, \cdots, S$}
		\State ${\cal B}_s \gets \left\{B\big(\cdot, \cdot, 2^s\big) \Big| B\text{ are not overlapping}, \exists (k,l) \in  B\ s.t.\ B_{k,l}= 1\right\}$
		\State $N_s \gets \text{ the cardinal number of }{\cal B}_s \text{ , i.e., card}\left({\cal B}_s\right)$
		\State Populate ${\cal S}_I$ using the $s$ value		
		\State Populate ${\cal N}_I$ using the $N_s$ value
	\EndParFor
	\LineComment{\emph{Calculate $\text{HFD}_I$ using the least square regression}}
	\State Choose a linear regression model: $M_R \sim D^* \cdot + h^*$
	\State Choose a least square regression operator: $R_\text{LS}$	
	\State Apply the $R_\text{LS}$ operator to $\Big(M_R, \log\left({\cal N}_I\right), \log\left(1/2^{{\cal S}_I}\right)\Big)$ to get $D$
	\State $\text{HFD}_I  \gets D$
\EndProcedure
\end{algorithmic}
\end{algorithm}

\section{The Hausdorff Fractal Dimension (HFD)}
\label{sec_Hausdorff_Fractal_Dimensio_HFD}
Let's consider a binary image $I$. Then, the $S$-tuple of possible box sizes, ${\cal S}_I$, is defined as the integer interval $[0, S]$, where $2^{S} \geq \max \left(m, n\right)$: ${\cal S}_I=\left(s\right)_{s=0}^{S}$.\footnote{The lower limit of $s=0$ could be pushed down toward lower values using super resolution enhancement of the image $I$.} For every box size $2^s$ with an $s$ value from ${\cal S}_I$, the number of `non-overlapping' boxes with at least one object-pixel is denoted $N_s$. The $S$-tuple of all $N_s$ is denoted as ${\cal N}_I=\left(N_s\right)_{s=0}^{S}$. Starting with the definition of the Haussdorf Fractal Dimension (HFD):
\begin{equation}
D_I = \text{HFD}_I = \lim_{2^s\rightarrow 0} \frac{\log \left(N_s\right)}{\log{\left(1/2^s\right)}},
\label{eq_HFD_1}
\end{equation}
and because of the discrete and finite nature of the ${\cal S}_I$ and ${\cal N}_I$, an extrapolation could be instead used:
\begin{eqnarray}
\label{eq_HFD_disc_1}
D_I = \text{HFD}_I & = & \argmin_{D^*} R_{\text{LS}}\Big(\log \left(N_s\right) = D^* \log\left(1 / 2^s\right) + h^*\Big),\\
&& s.t.\quad N_s \in {\cal N}_I, s = 0, \cdots, S,\nonumber
\end{eqnarray}
where $R_{\text{LS}}$ stands for a Least Square regression operator. The complete algorithm to calculate the HFD is provide in Algorithm \ref{alg_HFD_1}.\footnote{An implementation can be found here: \url{http://www.mathworks.com/matlabcentral/fileexchange/30329-hausdorff--box-counting--fractal-dimension}.}

\begin{algorithm}[!htbp]
\small
\caption{Calculate the proposed MHFD.}
\label{alg_MHFD_1}
\begin{algorithmic}[1]
\Procedure{$\text{MHFD}_I = \mathbf{MHFDCalculate}$}{$I_{n,m}$}
	\LineComment{\emph{Optional preprocessing}}
	\State (Optional) $I \gets \text{Denoised } I$
	\State (Optional) $I \gets \text{Edges of } I$
	\State (Optional) $I \gets \text{Skeleton of } I$		
	\LineComment{\emph{Then, calculate the new $\widehat{\cal S}_I$ and $\widehat{\cal N}_I$}}
	\State $S \gets \max \left(\log_2 m, \log_2 n\right) + 1$
	\ParFor {$s = 0, \cdots, S$}
		\State $\widehat{\cal B}_s \gets \left\{B\big(\cdot, \cdot, 2^s\big) \Big| Bs\text{ are not overlapping}, \exists (k,l) \in  B\ s.t.\ B_{k,l}= 1\right\}$
		\State $\widehat{\cal B}_s \gets \left\{B \Big| B \in \widehat{\cal B}_s, \exists (k,l) \in  B\ s.t.\ B_{k,l}= 0\right\}$
		\State $\widehat{\cal B}_s \gets \left\{B \Big| B \in \widehat{\cal B}_s, U\big(0, \text{card}(B=1)+1\big)\sim {X}\ni x_B >1\right\}$
		\State $\widehat N_s \gets \text{ the cardinal number of }\widehat{\cal B}_s \text{ , i.e., card}\left(\widehat{\cal B}_s\right)$
		\State Populate $\widehat{\cal S}_I$ using the $s$ value		
		\State Populate $\widehat{\cal N}_I$ using the $\widehat N_s$ value
	\EndParFor
	\LineComment{\emph{Calculate $\text{MHFD}_I$ using the least square regression}}
	\State Choose a linear regression model: $M_R \sim D^* \cdot + h^*$
	\State Choose a least square regression operator: $R_\text{LS}$	
	\State Apply the $R_\text{LS}$ operator to $\Big(M_R, \log\left(\widehat{\cal N}_I\right), \log\left(1/2^{\widehat{\cal S}_I}\right)\Big)$ to get $\widehat D$
	\State $\widehat D \gets c \widehat D \text{ where } c \text{ is a normalization factor}$
	\State $\text{MHFD}_I  \gets \widehat D$
\EndProcedure
\end{algorithmic}
\end{algorithm}

In the next section, a modified HFD is presented.

\section{The Modified Hausdorff Fractal Dimension (MHFD)}
\label{sec_Modified_Hausdorff_Fractal_Dimension_MHFD}
As mentioned in the Introduction section, the input image $I$ may not completely satisfy the basic requirements of a fractal shape. To address this challenge, we introduce two features:
\begin{enumerate}
\litem{Explicit Valuation of Non-object Pixels} Only boxes that at least contain a $0$ pixel are counted in.
\litem{Probabilistic Discard of Noise Boxes} For boxes with a small number of $1$ pixels, there should be higher chance that they are excluded from the count.
\end{enumerate}
To impose these two features, the definition of $N_s$ is revised: For every $s$ and for every non-overlapping box $B$ of patch size $2^s$, the number of $1$ and $0$ pixels are calculated using the Integral Image representation \cite{Bradley2007,Shafait2008}. Let's assume these numbers are $n_{B,1}$ and $n_{B,0}$, respectively. If $n_{B,0}=0$, the box is immediately discarded. Otherwise, a dice of $n_{B,1}+1$ faces is rolled, and if the outcome is the face showing $1$, the box is discarded. Finally, for every $s$, only those boxes that have not been discarded are counted in to calculate the new $\widehat  N_s$. The MHFD can be calculated in a way similar to that of Equation (\ref{eq_HFD_disc_1}):
\begin{eqnarray}
\label{eq_MHFD_disc_1}
\widehat D_I = \text{MHFD}_I & = & \argmin_{D^*} R_{\text{LS}}\Big(\log \left(\widehat  N_s\right) = D^* \log\left(1 / 2^s\right) + h^*\Big),\\
&& s.t.\quad \widehat  N_s \in \widehat {\cal N}_I, s = 0, \cdots, S.\nonumber
\end{eqnarray}
The details of the MHFD calculations are provided in Algorithm \ref{alg_MHFD_1}.\footnote{A Matlab implementation has been provided here: \url{http://www.mathworks.com/matlabcentral/fileexchange/50790-modified-haussdorf-fractal-dimension}.} Please note that we also considered optional preprocessing, edge extraction, and skeleton extraction steps in the algorithm that could improve the performance. This is further discussed in Section \ref{sec_Illustrative_Examples}.

\section{The Illustrative Examples}
\label{sec_Illustrative_Examples}

\begin{figure}
\centering
\begin{tabular}{cc}
\fbox{\includegraphics[width=2.5in]{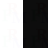}} &
\fbox{\includegraphics[width=2.5in]{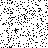}} \\
(a) & (b) \\
$\begin{array}{rcl}
\text{HFD} & = & 1.6809 \\
\text{MHFD} & = & 1.2876
\end{array}$
& 
$\begin{array}{rcl}
\text{HFD} & = & 1.8234 \\
\text{MHFD} & = & 1.3626
\end{array}$ \\
\fbox{\includegraphics[width=2.5in]{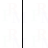}} &
\fbox{\includegraphics[width=2.5in]{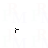}} \\
(c) & (d) \\
$\begin{array}{rcl}
\text{HFD} & = & 0.9259 \\
\text{MHFD} & = & 0.8242 
\end{array}$
& 
$\begin{array}{rcl}
\text{HFD} & = & 0.3952 \\
\text{MHFD} & = & 0.3101
\end{array}$ 
\end{tabular}
\caption{
a) and b) Two synthesized images with their associated HFDs and MHFDs.
c) and d) The preprocessed images of (a) and (b) and their associated HFDs and MHFDs.}
\label{fig_synt1}
\end{figure}

We start with two synthesized images, as shown in Figure \ref{fig_synt1}. Figure \ref{fig_synt1}(a) shows an image with a low level of complexity. Using the algorithm of Section \ref{sec_Hausdorff_Fractal_Dimensio_HFD}, a high HFD is obtained ($\text{HFD} = 1.6809$). In contrast, the MHFD gives a much lower dimension. If the edge of the image is used, which is equivalent to extracting the fractal shape, both HFD and MHFD provide low  dimensions around $0.9$ that better represent the complexity of the image. The second synthesized image is a binary salt-and-pepper noise (Figure \ref{fig_synt1}(b)). Again the HFD give a high dimension ($\text{HFD} = 1.8234$) while the MHFD is able to show more resiliency to the noise. When preprocessing and skeletonization are used (Figure \ref{fig_synt1}(d)), again both the HFD and MHFD provide a correct low dimension of $0.3$.

\begin{figure}
\centering
\begin{tabular}{ccc}
\fbox{\includegraphics[width=1.5in]{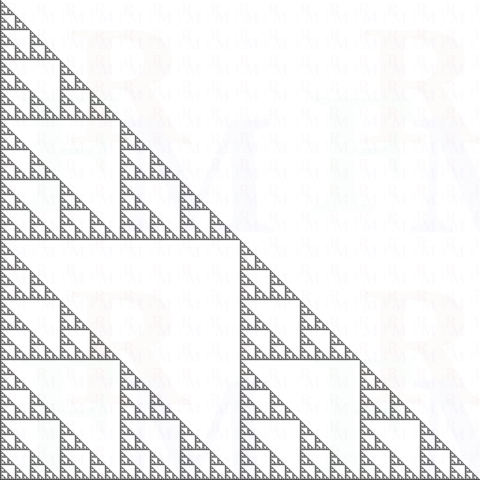}} &
\fbox{\includegraphics[width=1.5in]{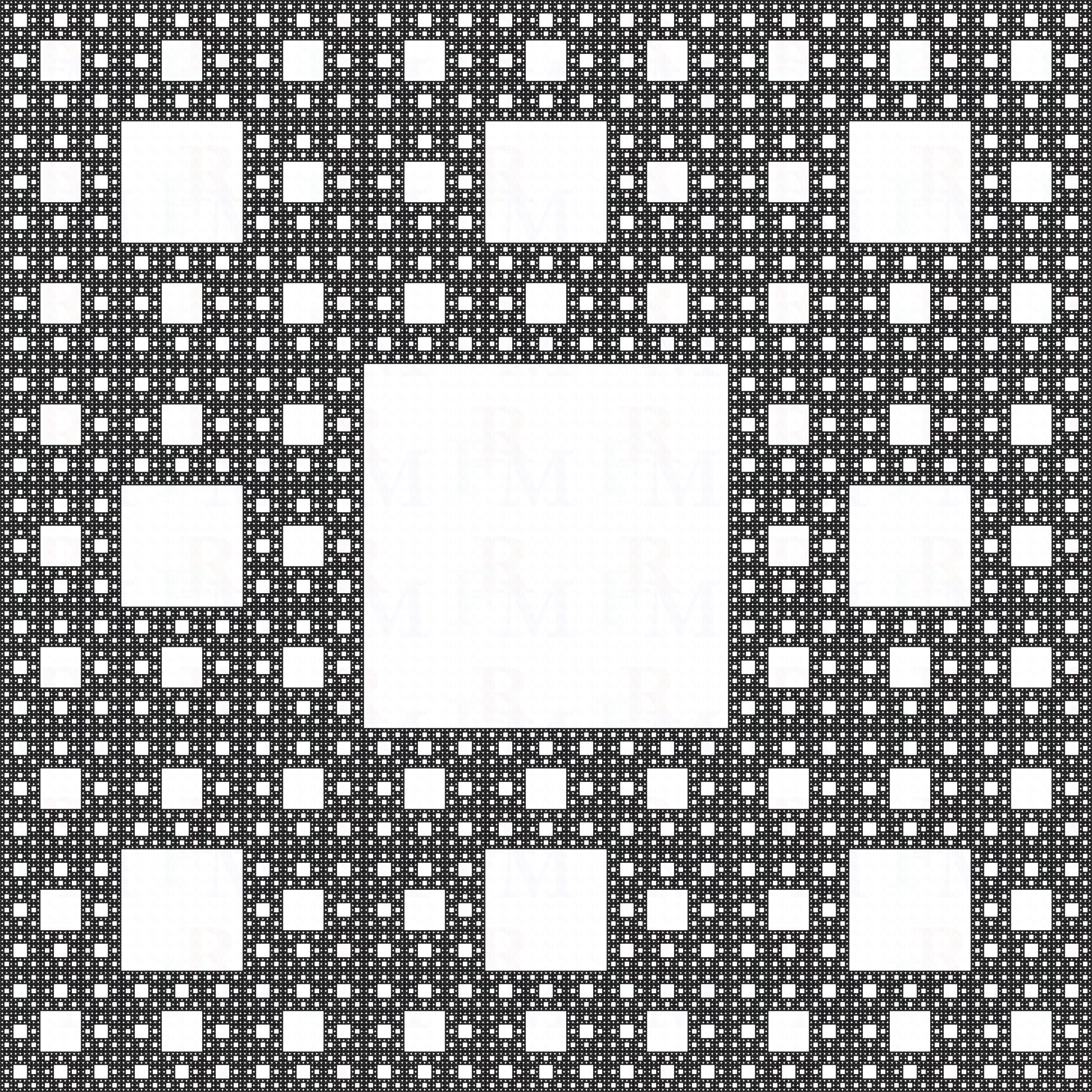}} &
\fbox{\includegraphics[width=1.5in]{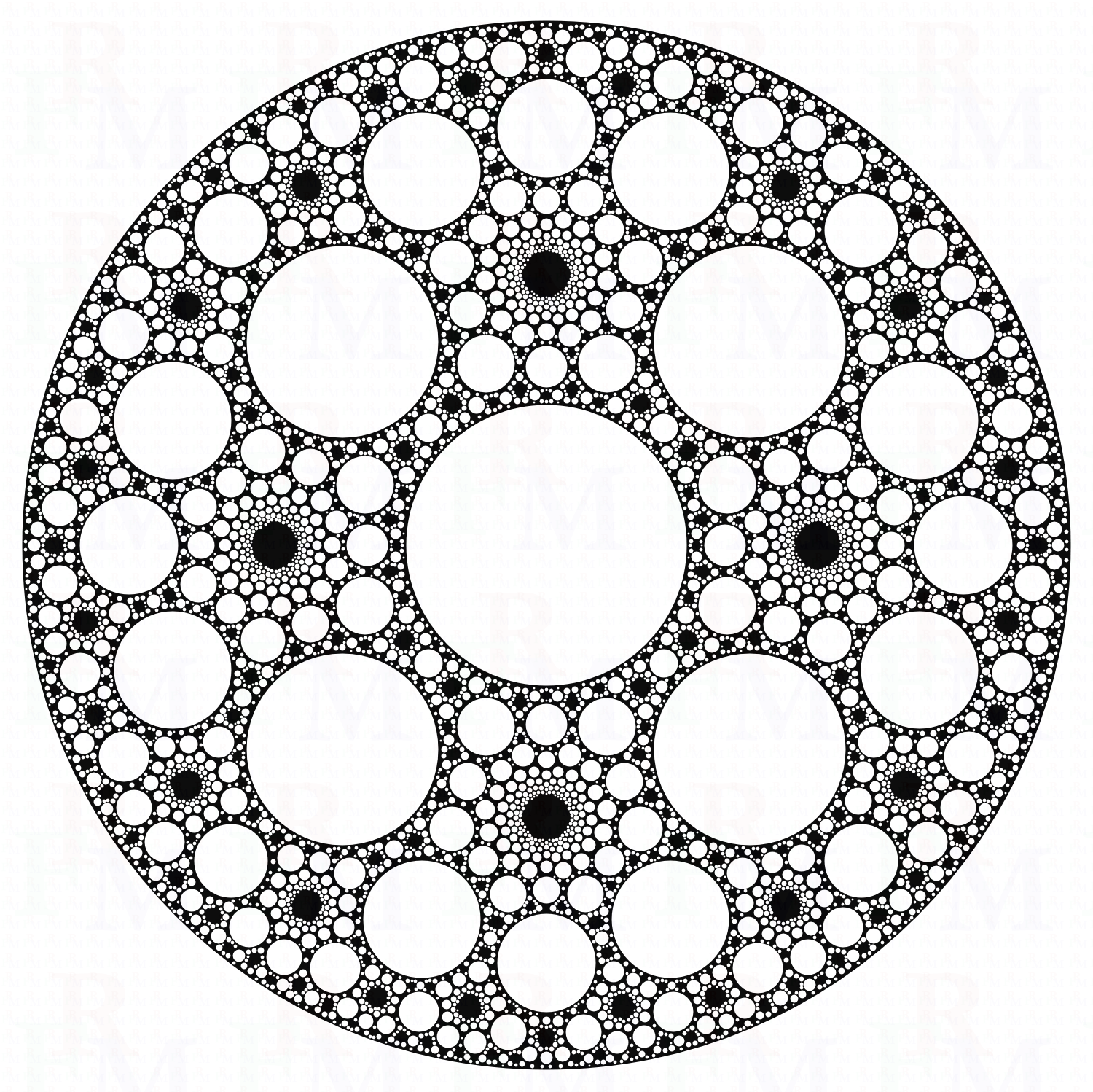}} \\
(a) Sample 1.$^\dagger$ & (b) Sample 2.$^\ddagger$ & (c) Sample 3.$^\mathsection$ \\
\fbox{\includegraphics[width=1.5in]{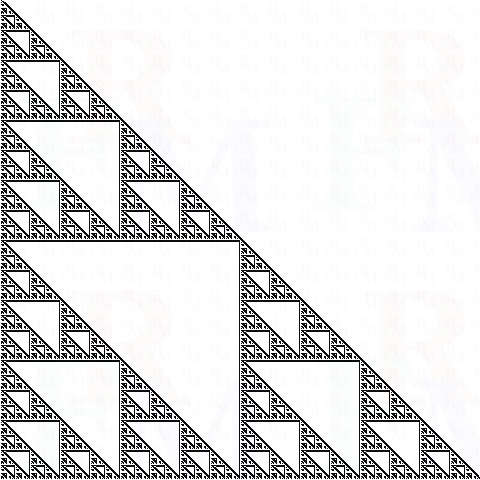}} &
\fbox{\includegraphics[width=1.5in]{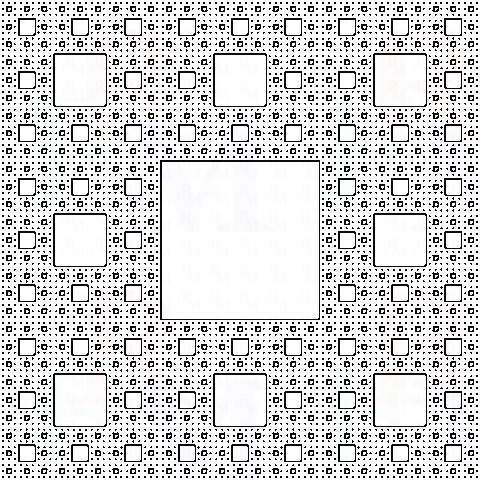}} &
\fbox{\includegraphics[width=1.5in]{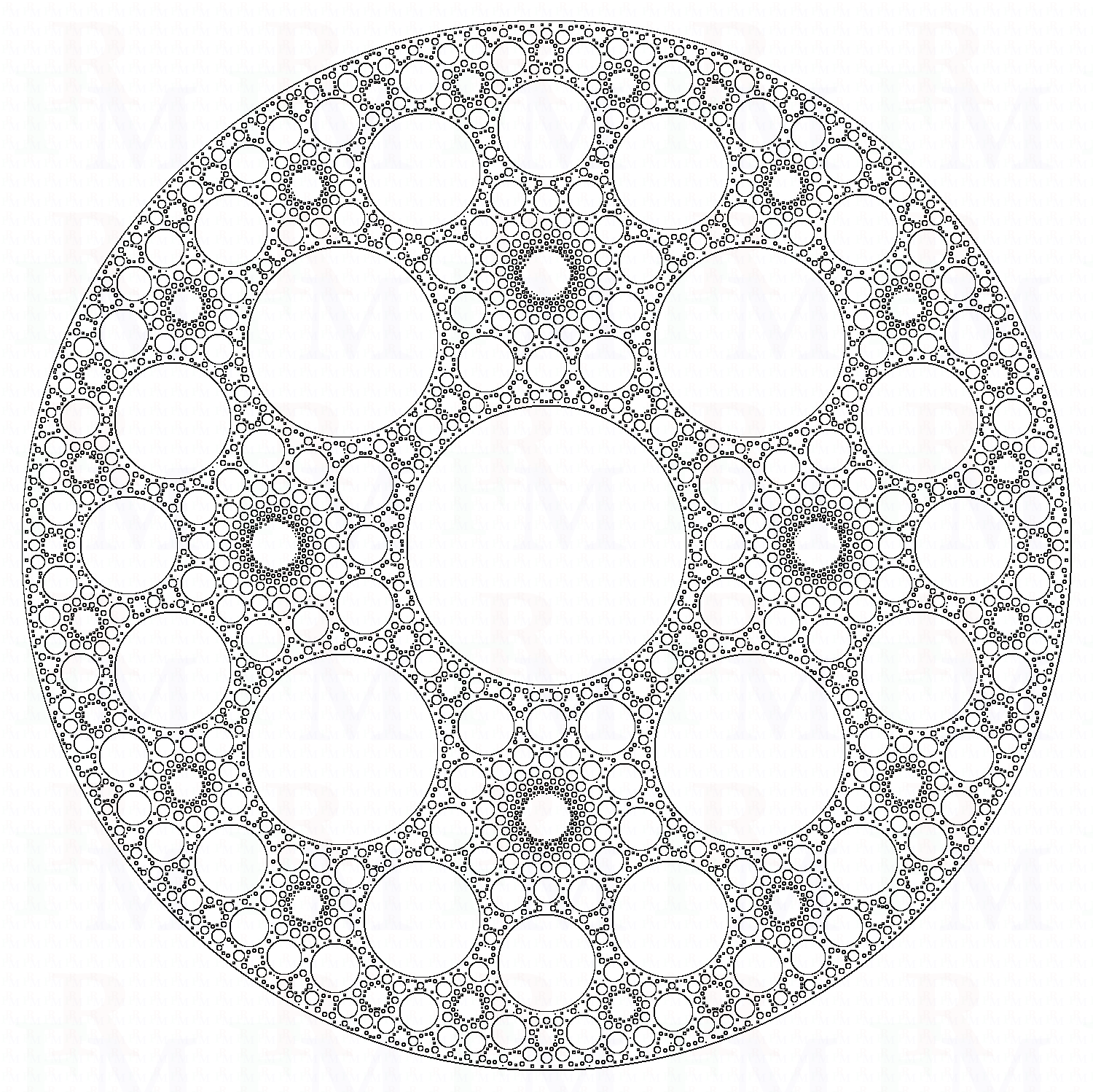}} \\
(d)  & (e)  & (f) \\
$\begin{array}{rcl}
\text{HFD} & = & 1.5999 \\
\text{MHFD} & = & 1.5843
\end{array}$
& 
$\begin{array}{rcl}
\text{HFD} & = & 1.8811 \\
\text{MHFD} & = & 1.6794
\end{array}$
& 
$\begin{array}{rcl}
\text{HFD} & = & 1.7129 \\
\text{MHFD} & = & 1.6263
\end{array}$ 
\end{tabular}
\caption{Three fractal images and their associated HFDs and MHFDs. a)-c) Input images.
d)-f) The edge images used to calculate the MHFD.
Notes: $^\dagger$ Link: \protect\url{http://4.bp.blogspot.com/-aHCfmDvyzFU/Un_U-Neo_GI/AAAAAAAAGpQ/DWzjztkh4HM/s1600/sierpinski.png}. 
$^\ddagger$ Link: \protect\url{http://upload.wikimedia.org/wikipedia/commons/thumb/a/a0/Sierpinski_carpet.png/480px-Sierpinski_carpet.png}.
$^\mathsection$ Link: \protect\url{http://www.math.upenn.edu/~pstorm/images/round_Sierpinski_carpet.png}.
}
\label{fig_fractals1}
\end{figure}

In the second example, some fractal images are considered. This examples are shown in Figure \ref{fig_fractals1}. It is worth mentioning that the normalization factor of Algorithm \ref{alg_MHFD_1} has been chosen in such a way that the MHFD of Figure \ref{fig_fractals1}(a) matches its analytic value. Also, in all cases, the edge extraction step has been used. Again, a better robustness can be observed in the values of the MHFD.

\begin{figure}
\centering
\small
\setlength{\tabcolsep}{1pt}
\begin{tabular}{cccc}
\twincludegraphics[height=1.0in]{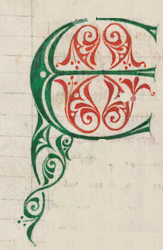} &
\twincludegraphics[height=1.0in]{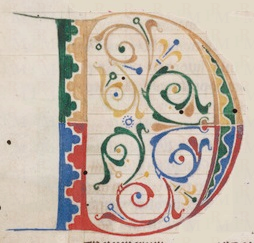} &
\twincludegraphics[height=1.0in]{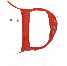} &
\twincludegraphics[height=1.0in]{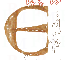} \\
(a) & (b) & (c) & (d) \\

\twincludegraphics[height=1.0in]{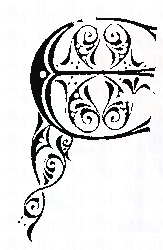} &
\twincludegraphics[height=1.0in]{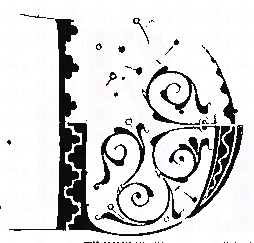} & 
\twincludegraphics[height=1.0in]{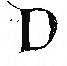}  &
\twincludegraphics[height=1.0in]{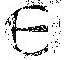} \\
(e) & (f) & (g) & (h) \\
$\begin{array}{rcl}
\text{HFD} & = & 1.5380 \\
\text{MHFD} & = & 1.4357
\end{array}$
& 
$\begin{array}{rcl}
\text{HFD} & = & 1.6135 \\
\text{MHFD} & = & 1.4930
\end{array}$
& 
$\begin{array}{rcl}
\text{HFD} & = & 1.3943 \\
\text{MHFD} & = & 1.2692
\end{array}$
& 
$\begin{array}{rcl}
\text{HFD} & = & 1.3983 \\
\text{MHFD} & = & 1.2111
\end{array}$\\
\twincludegraphics[height=1.0in]{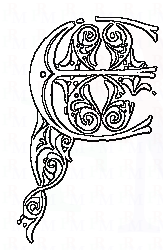} &
\twincludegraphics[height=1.0in]{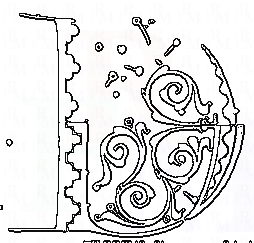} & 
\twincludegraphics[height=1.0in]{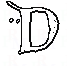}  &
\twincludegraphics[height=1.0in]{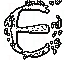} \\
(i) & j) & (k) & (l) \\
$\begin{array}{rcl}
\textbf{HFD} & = & \bf 1.5804 \\
\textbf{MHFD} & = & \bf 1.5078
\end{array}$
& 
$\begin{array}{rcl}
\textbf{HFD} & = & \bf 1.6091 \\
\textbf{MHFD} & = & \bf 1.5335
\end{array}$
& 
$\begin{array}{rcl}
\textbf{HFD} & = & \bf 1.5482 \\
\textbf{MHFD} & = & \bf 1.4472
\end{array}$
& 
$\begin{array}{rcl}
\textbf{HFD} & = & \bf 1.3863 \\
\textbf{MHFD} & = & \bf 1.2794
\end{array}$\\
\twincludegraphics[height=1.0in]{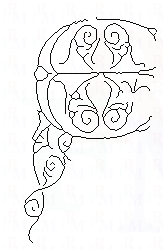} &
\twincludegraphics[height=1.0in]{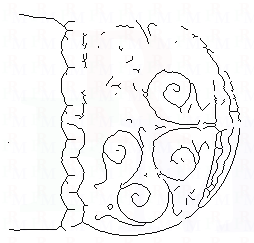} & 
\twincludegraphics[height=1.0in]{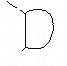}  &
\twincludegraphics[height=1.0in]{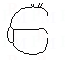} \\
(m) & (n) & (o) & (p) \\
$\begin{array}{rcl}
\text{HFD} & = & 1.3731 \\
\text{MHFD} & = & 1.2944
\end{array}$
& 
$\begin{array}{rcl}
\text{HFD} & = & 1.4069 \\
\text{MHFD} & = & 1.3248
\end{array}$
& 
$\begin{array}{rcl}
\text{HFD} & = & 1.1629 \\
\text{MHFD} & = & 1.0619
\end{array}$
& 
$\begin{array}{rcl}
\text{HFD} & = & 1.0864 \\
\text{MHFD} & = & 0.9594
\end{array}$
\end{tabular}
\caption{
a) and b) Samples of Type I ({\em Littera Notabilior}) objects.
c) and d) Samples of Type II (Enlarged Capital) objects.
e)-h) Simple binarization of (a) to (d) and their associated HFDs and MHFDs.
i)-l) The same as (e) to (h) but with edge extraction.
m)-p) Preprocessed and skeletonization of (a) to (d) and their associated HFDs and MHFDs.
}
\label{fig_hist1}
\end{figure}

As the final example, some illustrated alphabetical letters from historical manuscripts are considered.\footnote{More specifically, from a manuscript entitled ``Collectio decem partium (expansion of Ivo of Chartres, Panormia).'' More information at: \url{http://parkerweb.stanford.edu/parker/actions/manuscript_description_long_display.do?ms_no=94}.} In particular, Figures \ref{fig_hist1}(a) and \ref{fig_hist1}(b) are {\em Litterae Notabiliores}, i.e., enlarged letters within a text, designed to clarify the syntax of a passage. In contrast, two other examples, shown in Figures \ref{fig_hist1}(c) and \ref{fig_hist1}(d), are regular enlarged capitals. It could be easily observed that the {\em Littera Notabilior} class shows a higher level of complexity compared to the enlarged capital class. Therefore, it could be argued that a fractal dimension analysis would be capable to differentiate between these two classes. Three alternative cases are considered: i) A simple thresholding algorithm \cite{Otsu1979} is applied to the input images and then the HFD and MHFD are calculated (Figures \ref{fig_hist1}(e)-(h)), ii) the edges of the binary images are used in the calculations (Figures \ref{fig_hist1}(i)-(l)), and iii) the color images are converted to gray using the min-average color-to-gray transform \cite{Farrahi2010} and then the skeleton image is used in the calculations (Figures \ref{fig_hist1}(m)-(p)).
As can be seen from the values calculated using the HFD and MHFD, the MHFD provides a better robustness. In particular, for the edge images, i.e., Figures \ref{fig_hist1}(i)-(l), the average intra-class distance is reduced from $0.095$ to $0.075$ when switching from the HFD to the MHFD, while the inter-class distance is increased from $0.125$ to $0.155$, respectively. This could be interpreted as an increase of 57\% in the class differentiability if the MHFD is used, which could be crucial for images that fall in the border of the two classes. We will evaluate the performance of the measure using a bigger dataset in the future.

\begin{figure}
\centering
\small
\setlength{\tabcolsep}{2pt}
\begin{tabular}{cc}
\includegraphics[height=2.0in]{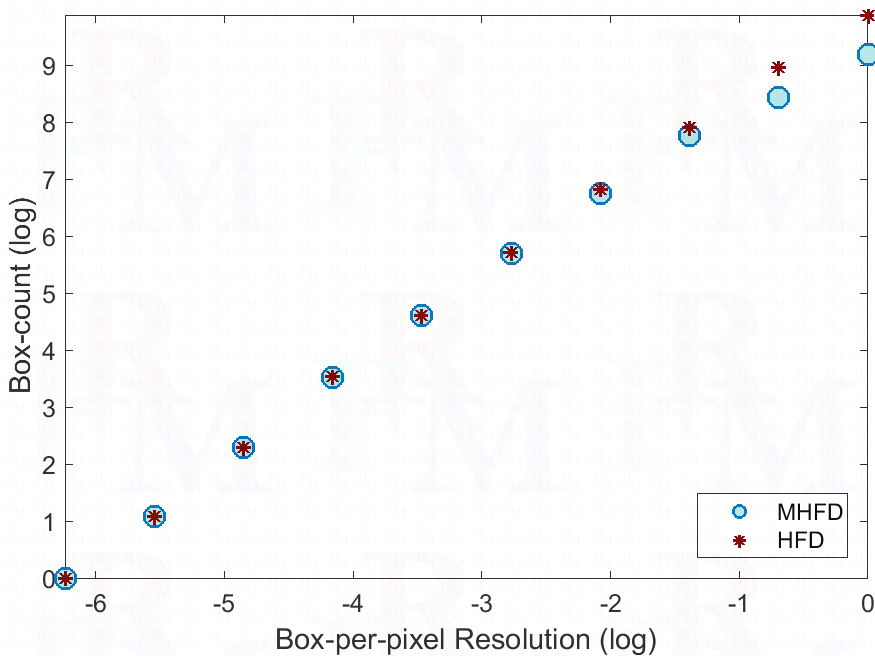} &
\includegraphics[height=2.0in]{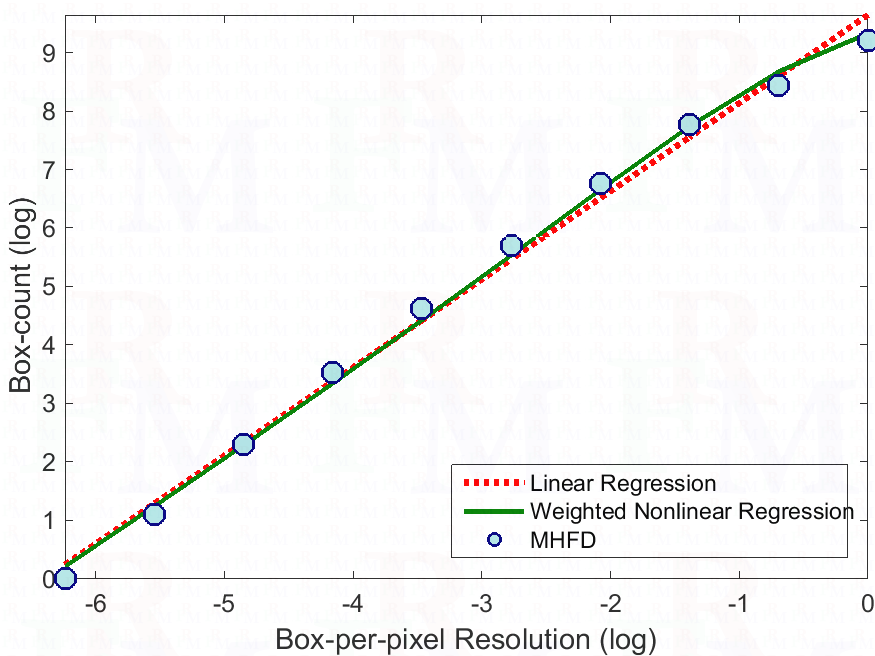} \\
(a) & (b) 
\end{tabular}
\caption{
a) The behavior of the discrete ${\cal S}_I$ and ${\cal N}_I$ associated to the HFD and also that of $\widehat{\cal S}_I$ and $\widehat{\cal N}_I$ associated to the MHFD. The values correspond to the image shown in Figure \protect\ref{fig_fractals1}(a). The deflection introduced by the MHFD for high values of the high box-per-pixel resolution can be easily observed.
b) A comparison between linear regression of Equation (\protect\ref{eq_HFD_disc_1}) and weighted nonlinear regression of Equation (\protect\ref{eq_MHFD_disc_2}) in fitting the MHFD data. The nonlinear model provides a better fit especially for the high box-per-pixel resolutions. 
}
\label{fig_Reg1}
\end{figure}

Finally, we would like to present a preliminary analysis of the other impacts of the MHFD. It can be easily argued that even for true fractal shapes the MHFD discards a high ratio of boxes for $s$ values near 1, which correspond to high values of the box-per-pixel resolutions, i.e., $1/2^s$. To be specific, for single-pixel boxes, which correspond to $s=0$ and a box-per-pixel resolution of 1, 50\% of potential boxes are on average discarded. A less sever effect is expected for lower resolutions. This phenomenon, which can be imagined as a `deflection' effect and shown in Figure \ref{fig_Reg1}, would weaken applicability of the Hausdorff model (\ref{eq_HFD_1}). To contain the impact of deflection, we propose to use a modified nonlinear\footnote{in the sense of the log values of $\widehat{\cal S}_I$ and $\widehat{\cal N}_I$.} model as follows:
\begin{eqnarray}
\label{eq_MHFD_disc_2}
\widehat D_I = \text{MHFD}_I & = & \argmin_{D^*} R_{\text{LS}}\Big(\log \left(\widehat  N_s\right) = \frac{D^* \log\left(1 / 2^s\right)}{1 / 2^s + 0.9}  + h^*, W_s\Big),\\
&& s.t.\quad 
\left\{\begin{array}{rcl}
\widehat  N_s & \in & \widehat {\cal N}_I, s = 0, \cdots, S,\nonumber\\
W_s & = & \Big(1/ 2^s\Big)_{s=0}^S,\end{array}\right.
\end{eqnarray}
where $W_s$ is the weight vector to be considered in the least square minimization. 
Figure \ref{fig_Reg1}(b) shows the performance of the proposed weighted nonlinear regression. The green curve, which corresponds to Equation (\ref{eq_MHFD_disc_2}), provides a better fit especially for low $s$  (high box-per-pixel resolution) values. We further investigate this aspect in the future.

\section{The Conclusions}
\label{sec_Conclusions}
A modified box-counting Hausdorff fractal dimension has been introduced in order to apply it in complexity analysis of binary images. The core of the modification is based on two features that weaken the requirement of presence of a shape and at the same time discard potential noise boxes in a probabilistic way. In addition, preprocessing of the input images along with edge or skeleton extraction has been considered. The proposed method has been tested on noise images, fractal shapes, and also illustrated manuscript images. In all cases, the modified dimension showed robustness even in absence of edge or skeleton extraction steps. 

In the future, the performance on a bigger dataset of manuscript images will be evaluated. In addition, irregular pooling of boxes to weaken the non-overlapping consideration toward a more robust estimation will be considered. Finally, nonlinear regression models will be considered in order to better fit and absorb the deflection introduced by the proposed method in small box sizes.

\section*{References}

\bibliographystyle{plain} \bibliography{imagep}

\end{document}